\documentclass{article}

\PassOptionsToPackage{numbers, compress}{natbib}

\usepackage[square,numbers]{natbib}
\usepackage[final]{neurips_2022}

\usepackage[utf8]{inputenc} 
\usepackage[T1]{fontenc}    
\usepackage{hyperref}       
\usepackage{url}            
\usepackage{booktabs}       
\usepackage{amsfonts}       
\usepackage{nicefrac}       
\usepackage{microtype}      
\usepackage{xcolor}          
\usepackage{graphicx}
\usepackage{natbib}
\usepackage{gensymb}
\usepackage{xcolor}        
\graphicspath{ {./img/} }

\title{Real-time Health Monitoring of Heat Exchangers using Hypernetworks and PINNs}

\author{%
  Ritam Majumdar\\
  TCS Research\\
  \texttt{ritam.majumdar@tcs.com} \\
  \And
  Vishal Jadhav \\
  TCS Research \\
  \texttt{vi.suja@tcs.com} \\
  \AND
  Anirudh Deodhar \\
  TCS Research \\
  \texttt{anirudh.deodhar@tcs.com} \\
  \And
  Shirish Karande \\
  TCS Research\\
  \texttt{shirish.karande@tcs.com} \\
  \And
  Lovekesh Vig \\
  TCS Research \\
  \texttt{lovekesh.vig@tcs.com} \\
  \And
  Venkataramana Runkana \\
  TCS Research \\
  \texttt{venkat.runkana@tcs.com} \\
}

\begin{document}
	
\maketitle

\begin{abstract}
 
 We demonstrate a Physics-informed Neural Network (PINN) based model for real-time health monitoring of a heat exchanger, that plays a critical role in improving energy efficiency of thermal power plants. A hypernetwork based approach is used to enable the domain-decomposed PINN learn the thermal behavior of the heat exchanger in response to dynamic boundary conditions, eliminating the need to re-train. As a result, we achieve orders of magnitude reduction in inference time in comparison to existing PINNs, while maintaining the accuracy on par with the physics-based simulations. This makes the approach very attractive for predictive maintenance of the heat exchanger in digital twin environments.

\end{abstract}

\section{Introduction}
\label{sec:intro}

Air preheaters (APH) are regenerative heat exchangers used in thermal power plants, that improve the boiler thermal efficiency by up to 10\%, cutting the greenhouse gas emissions by up to a million tons annually. Fouling is a major issue faced by most APH, that causes forced outages of entire plant incurring huge revenue losses. Fouling is governed by a complex multi-physics phenomena among interacting gases and metals (see schematic of APH in Figure \ref{fig:CompDomain}). Absence of sensors within the APH makes it difficult to monitor its condition, complicating the on-field decision-making. Accurate estimation of internal thermal profile as a function of external sensor measurements can go a long way in monitoring and controlling this fouling.

There have been several efforts towards this using physics-based modeling \cite{WANG201952}, data-driven modeling \cite{Gupta2021} and more recently physics-informed neural networks (PINN) \cite{Jadhav2022}. The case for utility of PINNs is already well established given their superior inference speed and reduced dependence on sensor data for training \cite{RAISSI2019686}. However, digital twins \cite{Digitaltwin} of industrial systems such as APH need real-time inference with dynamically changing conditions; a challenge PINNs fail to address \cite{WANG2022114424}. PINNs require re-training when boundary conditions change, which involves large cost and effort. Several approaches including transfer learning \cite{CHAKRABORTY2021109942,Desai2021}, meta learning \cite{Penwarden2021}, a mosaic flow predictor \cite{WANG2022114424} and HyperPINNs \cite{Filipe2021} have been attempted. However, most of these solutions either do not eliminate the re-training requirement completely or have been demonstrated on very simplified problems with simplified physics/geometry. With a multi-sector arrangement inducing discontinuity in thermal profile, multiple fluids interacting with each other through a rotating porous solid material and a non-standard geometry, real-time inference for APH through PINNs is quite challenging.

In this work, we demonstrate a hypernetwork based PINN solution for training-free inference enabling real-time internal temperature estimation. The hypernetwork is trained based on weights extracted from PINNs trained on multiple tasks (a designed set of varying boundary conditions). A case of using Taguchi design of experiments \cite{taguchi} for identifying the optimal trade-off between accuracy and training cost is also presented. The hypernetwork predicts the weights for an arbitrary task (boundary condition) in near real-time without a significant loss of accuracy vis-a-vis physics-based simulation. To the best of our knowledge, this work is the first demonstration of a PINN model trained to predict for unseen boundary conditions for a complex industrial scale heat exchanger.

\section{Methods}
\label{sec:methods}

Heat transfer in APH can be described as conduction and convection phenomena occurring between gases, air and rotating metal packing across multiple sectors and is governed by non-dimensional system of equations and boundary conditions \cite{SKIEPKO19882227}. For details of the equations see appendix~\ref{app:domaindecomPINN}. 

We replicate a domain-decomposed PINN for APH proposed by \cite{Jadhav2022} and use it a base PINN in this work. Their PINN $u_\Theta (\bf x)$ is a union of multiple individual PINNs $ u_{\Theta_j} (\bf x) $ for each fluid medium j ($ \forall j =  1,2,3 $) woven together by specific interface conditions (eqns.~\ref{eq:intface_gas},\ref{eq:intface_priair},\ref{eq:intface_secair}). The complete solution is given as:
\begin{equation}
    u_\Theta (\bf x) = {\bigcup_{j=1}^{3}} u_{\Theta_j} (\bf x)
\end{equation}
However, this base PINN is difficult to generalize for dynamic boundary conditions and therefore needs re-training if and when the conditions change, incurring computational costs and additional time. To enable training-free inference on these PINNs, we train a hypernetwork \cite{David2016Hypernetworks} for predicting the weights of the base PINN, as a function of parameterized dynamic conditions/tasks ($\lambda$). It is represented by equation \ref{eq:hypernetwork}, where $\Theta_h$ represents the learnable parameters for the hypernetwork and $\Theta$ represents the weights of the PINN corresponding to parametric conditions $\lambda$.
\begin{equation}
    \label{eq:hypernetwork}
    \Theta = N_{\Theta_h}(\lambda)
\end{equation}
\begin{figure}
    \centering
    \includegraphics[width=\textwidth]{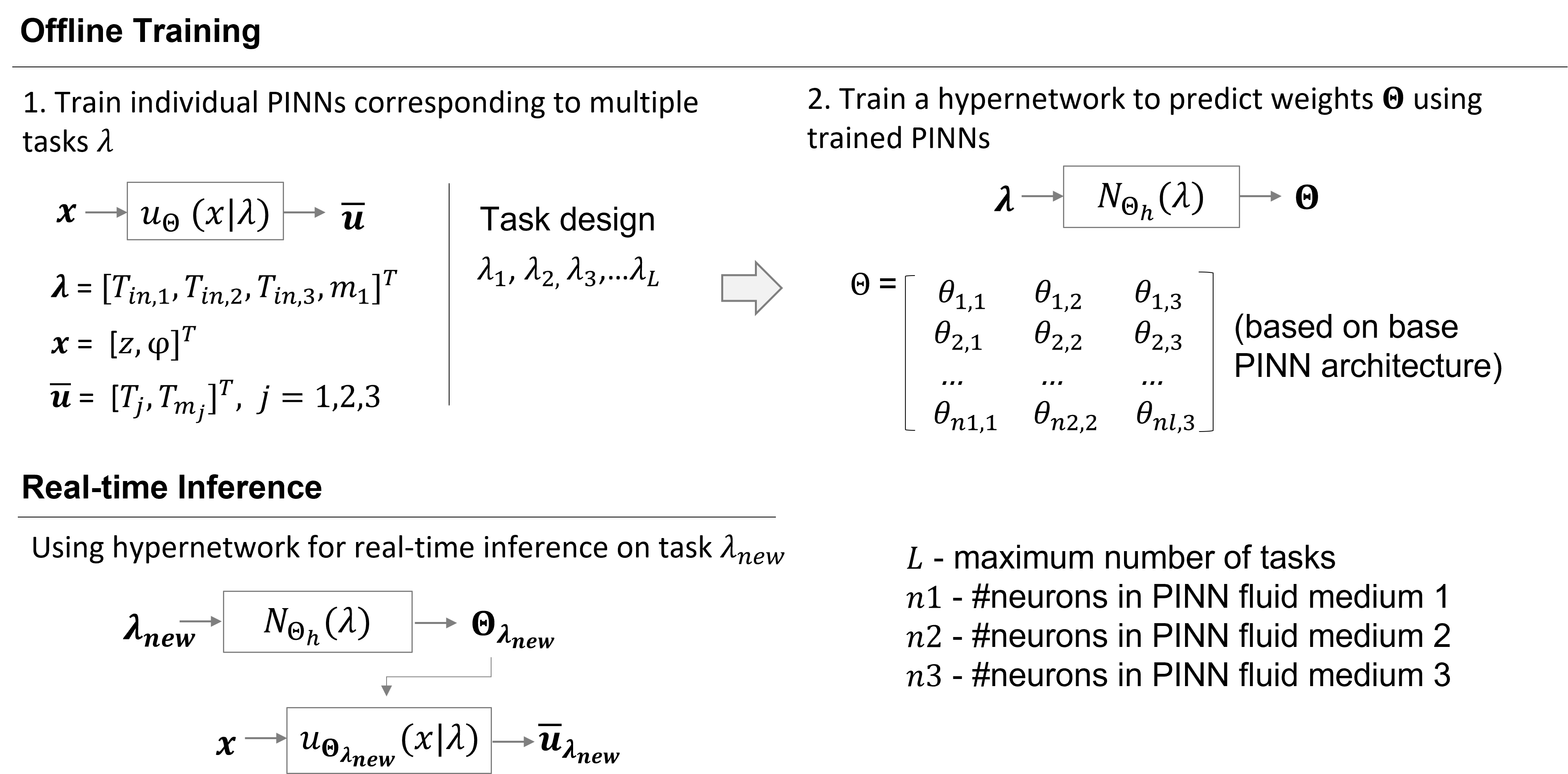}
    \caption{Training and Inference - Hypernetwork for Domain Decomposed PINN}
    \label{fig:methodology}
\end{figure}

The training data for the hypernetwork is generated by creating several base PINN models for varying boundary conditions ($\lambda$) over a typical range observed in the field. The boundary conditions include inlet temperature and flow-rate conditions for APH. A hypernetwork is then trained using the weights from this base PINN set, as shown in figure \ref{fig:methodology}. The idea is to enable the hypernetwork to capture the dynamics introduced due to changing boundary values based on a bank of pre-generated PINN solutions. The rationale behind learning weights of base PINN instead of directly the thermal field is the flexibility and adaptability it provides. In industrial settings, when measurements may come at different frequency and granularity, this PINN-based hypernetwork can provide a unified way to integrate and learn from all such data. It would be tough to do the same if the thermal field values are directly used in the hypernetwork. Once trained, this hypernetwork and base PINN together enable real-time inference of the thermal field as a function of previously unseen boundary conditions (available as sensor values on field). The hypernetwork architecture and hyper-parameter details are provided in appendix \ref{app:hypernetdetails}. 
\section{Results and Discussion}
\label{sec:ResultsandDiscussion}
\begin{figure}
    \centering
    \includegraphics[width=\textwidth]{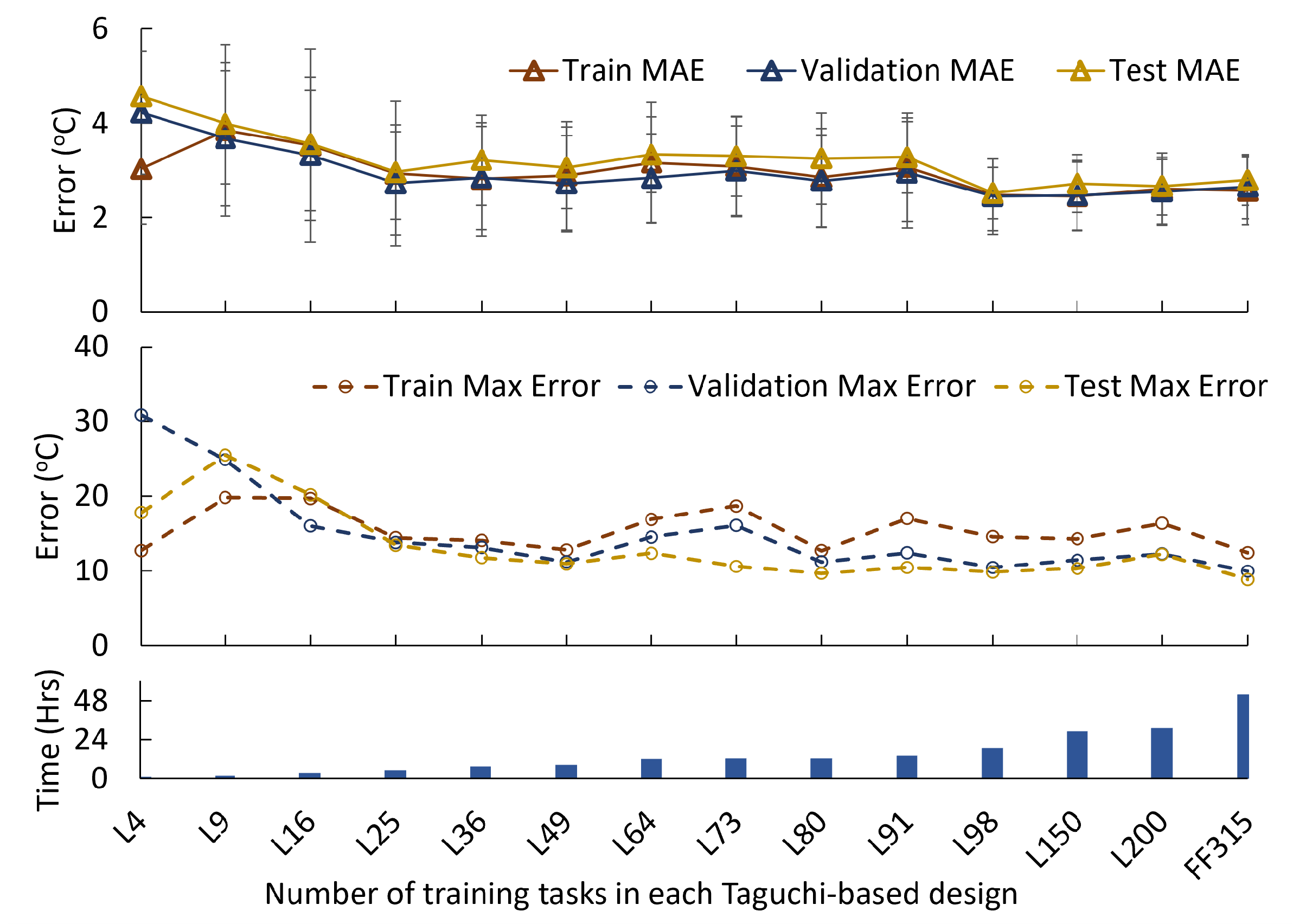}
    \caption{Mean absolute and Max error for different design of training tasks for hypernetwork PINNs and total time required for training the hypernetwork for each design}
    \label{fig:HypernetworkComparison}
\end{figure}

To observe the training cost and accuracy trade-off, base PINNs were trained for all combinations of tasks (Full factorial design FF with 315 tasks) within the given ranges (table \ref{tab:variablerange}). Hypernetworks were then trained by creating different designs (L4-L200) of base PINN tasks by varying the step size, inspired by Taguchi design of experiments \cite{taguchi}. For instance, L200 indicates a training set of 200 different tasks distributed across the search space and orthogonal to each other (see appendix \ref{app:Taguchi}. 
The hypernetwork approach is evaluated based on the accuracy of inference, the training time/effort and the inference time. First, the inference from hypernetwork based PINN is compared against that from a physics-based model, that uses a finite difference method to solve the set of governing equations, recreated and validated based on \cite{WANG201952}. As seen in Figure \ref{fig:HypernetworkComparison}, Mean Absolute Error (MAE) across all the train, validation and test sets (see table \ref{tab:validtesttask}) is under 4 \degree C, indicating that hypernetwork based PINNs are able to predict at a satisfactory level of accuracy. However, the maximum error only improves once the number of training tasks increases with each design, also increasing the training costs. A trade-off between the desired accuracy and the required training costs can be achieved based on Figure \ref{fig:HypernetworkComparison}. For example, while the accuracy over task designs L49 and FF315 is within the acceptable range, the offline training time increases by almost 500\% for FF315 design. Depending upon the constraints, a suitable design of tasks can be selected.

\begin{figure}
    \centering
    \includegraphics[width=\textwidth]{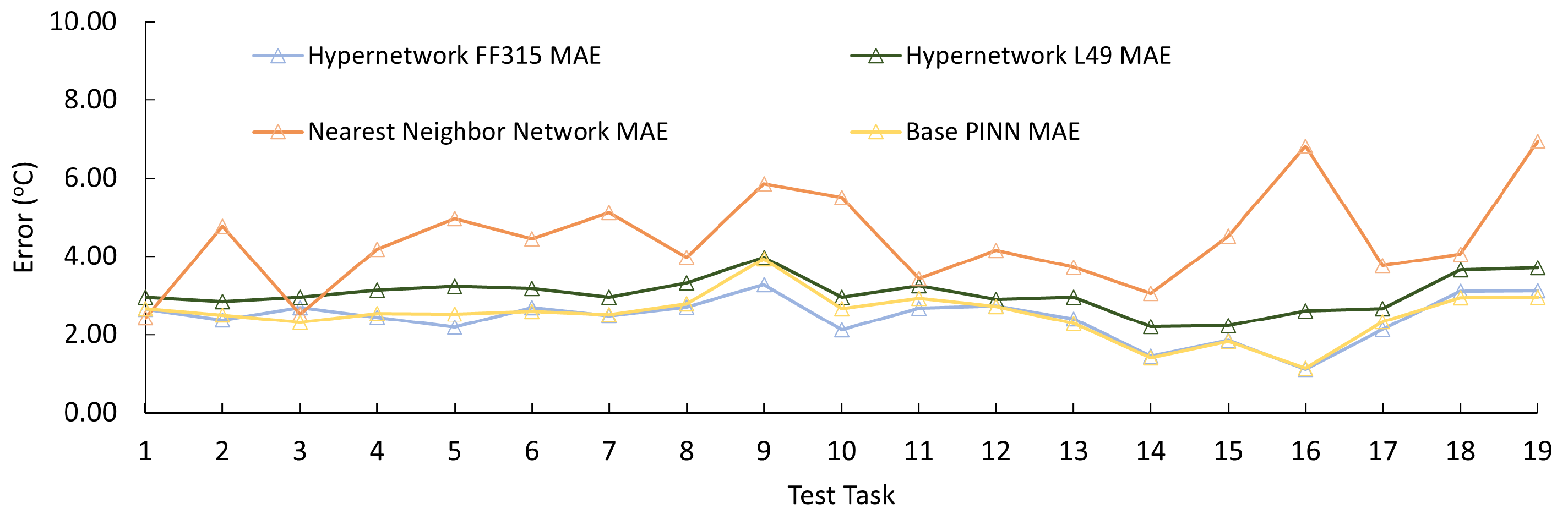}
    \caption{Comparison of test set errors for Hypernetwork trained on L49 task design, FF315 task design, base PINN approach and a naive nearest-neighbor approach based on FF315 task design}
    \label{fig:NearestNeighborComparison}
\end{figure}

Figure \ref{fig:NearestNeighborComparison} shows the comparison of MAE on each of the tasks in the test set with hypernetworks trained using L49 and FF315 design against the base PINN \cite{Jadhav2022} and the naive method. The naive method here refers to a simplified method of selecting the nearest (neighbor) available PINN for inferring the thermal field for a given condition (without training a hypernetwork). Both the base PINN and the hypernetwork-based PINNs outperform the naive inference consistently. Often, small perturbations to the external condition lead to significant variations in the thermal field inside the APH, which the naive approach may sometimes miss that PINNs seem to capture.The accuracy of hypernetwork based PINNs appears to match the base PINN accuracy almost across all the test tasks. 

However, the major advantage the hypernetwork based PINNs provide over the base PINNs is the training-free inference, as base PINNs need re-training for every new boundary condition. The inference time of hypernetwork based PINN ($\sim$5 sec) is quite superior both to the benchmark physics-based iterative solver ($\sim$200 sec) \cite{WANG201952} and the transfer-learnt base PINN ($\sim$180 sec) \cite{Jadhav2022}. 
Fouling within APH is a strong function of its internal thermal field, which governs the chemical reactions and location of deposition. An accuracy band of up to 5\degree C and near real-time inference is highly desirable for monitoring of fouling. The real-time accurate estimation of internal temperatures as a function of external sensor values (boundary conditions) reported here is significant, as it can be used for real-time estimation of fouling propensity \cite{Jadhav2022} as well as forecasting future progression of fouling \cite{Gupta2021}, to avoid forced outages of the plant.
\section{Conclusion}
\label{sec:conclusion}
A hypernetwork-based training regime for domain-decomposed PINNs \cite{Jadhav2022} is demonstrated, that enables the PINNs to learn the underlying physical behavior of the system without explicit re-training for dynamic boundary conditions. As a result, the hypernetwork-based PINN provides a significant improvement over iterative physical simulations or conventional PINNs in terms of inference time, without significant loss of accuracy. This model can enable real-time monitoring and control of APH fouling in an industrial digital twin, potentially avoiding huge revenue losses. To the best of our knowledge, this is the first demonstration of hypernetwork-based domain-decomposed PINNs successfully applied to a complex industrial heat exchanger. We plan to improve and extend the proposed methodology to accommodate changes in boundary type as well as learning from a combination of actual data and PINNs, which is critical for digital twin environment.

  
\section{Broad Impact}
Although, PINNs have been popular for scientific discovery, their industrial use has been limited due to boundary condition management issue among others. The methodology demonstrated here is generic and will make PINNs more amenable to industrial digital twins by reducing the inference time as well as doing away with re-training requirement. 





\bibliography{neurips_2022.bib}

\begin{thebibliography}{10}

\bibitem{WANG201952}
Limin Wang, Yufan Bu, Dechao Li, Chunli Tang, and Defu Che.
\newblock Single and multi-objective optimizations of rotary regenerative air
  preheater for coal-fired power plant considering the ammonium bisulfate
  deposition.
\newblock {\em International Journal of Thermal Sciences}, 136:52--59, 2019.

\bibitem{Gupta2021}
Ashit Gupta, Vishal Jadhav, Mukul Patil, Anirudh Deodhar, and Venkataramana
  Runkana.
\newblock {Forecasting of Fouling in Air Pre-Heaters Through Deep Learning}.
\newblock ASME 2021 Power Conference, 07 2021.
\newblock V001T01A002.

\bibitem{Jadhav2022}
Vishal Jadhav, Anirudh Deodhar, Ashit Gupta, and Venkataramana Runkana.
\newblock Physics informed neural network for health monitoring of an air
  preheater.
\newblock PHM Society European Conference, 7(1), 07 2022.

\bibitem{RAISSI2019686}
M.~Raissi, P.~Perdikaris, and G.E. Karniadakis.
\newblock Physics-informed neural networks: A deep learning framework for
  solving forward and inverse problems involving nonlinear partial differential
  equations.
\newblock {\em Journal of Computational Physics}, 378:686--707, 2019.

\bibitem{Digitaltwin}
Eric VanDerHorn and Sankaran Mahadevan.
\newblock Digital twin: Generalization, characterization and implementation.
\newblock {\em Decision Support Systems}, 145:113524, 2021.

\bibitem{WANG2022114424}
Hengjie Wang, Robert Planas, Aparna Chandramowlishwaran, and Ramin Bostanabad.
\newblock Mosaic flows: A transferable deep learning framework for solving pdes
  on unseen domains.
\newblock {\em Computer Methods in Applied Mechanics and Engineering},
  389:114424, 2022.

\bibitem{CHAKRABORTY2021109942}
Souvik Chakraborty.
\newblock Transfer learning based multi-fidelity physics informed deep neural
  network.
\newblock {\em Journal of Computational Physics}, 426:109942, 2021.

\bibitem{Desai2021}
Shaan Desai, Marios Mattheakis, Hayden Joy, Pavlos Protopapas, and Stephen
  Roberts.
\newblock One-shot transfer learning of physics-informed neural networks, 2021.

\bibitem{Penwarden2021}
Michael Penwarden, Shandian Zhe, Akil Narayan, and Robert~M. Kirby.
\newblock Physics-informed neural networks (pinns) for parameterized pdes: A
  metalearning approach, 2021.

\bibitem{Filipe2021}
Filipe de~Avila Belbute-Peres, Yi-fan Chen, and Fei Sha.
\newblock Hyperpinn: Learning parameterized differential equations with
  physics-informed hypernetworks, 2021.

\bibitem{taguchi}
Genichi Taguchi, Subir Chowdhury, and Yuin Wu.
\newblock {\em Taguchi's quality engineering handbook}.
\newblock John Wiley \& Sons, 2004.

\bibitem{SKIEPKO19882227}
T.~Skiepko.
\newblock The effect of matrix longitudinal heat conduction on the temperature
  fields in the rotary heat exchanger.
\newblock {\em International Journal of Heat and Mass Transfer},
  31(11):2227--2238, 1988.

\bibitem{David2016Hypernetworks}
David Ha, Andrew Dai, and Quoc~V. Le.
\newblock Hypernetworks, 2016.

\bibitem{Chung1983}
Chung-Hsiung Li.
\newblock {A Numerical Finite Difference Method for Performance Evaluation of a
  Periodic-Flow Heat Exchanger}.
\newblock {\em Journal of Heat Transfer}, 105(3):611--617, 08 1983.

\bibitem{tensorflow2015-whitepaper}
Mart\'{i}n Abadi, Ashish Agarwal, Paul Barham, Eugene Brevdo, Zhifeng Chen,
  Craig Citro, Greg~S. Corrado, Andy Davis, Jeffrey Dean, Matthieu Devin,
  Sanjay Ghemawat, Ian Goodfellow, Andrew Harp, Geoffrey Irving, Michael Isard,
  Yangqing Jia, Rafal Jozefowicz, Lukasz Kaiser, Manjunath Kudlur, Josh
  Levenberg, Dandelion Man\'{e}, Rajat Monga, Sherry Moore, Derek Murray, Chris
  Olah, Mike Schuster, Jonathon Shlens, Benoit Steiner, Ilya Sutskever, Kunal
  Talwar, Paul Tucker, Vincent Vanhoucke, Vijay Vasudevan, Fernanda Vi\'{e}gas,
  Oriol Vinyals, Pete Warden, Martin Wattenberg, Martin Wicke, Yuan Yu, and
  Xiaoqiang Zheng.
\newblock {TensorFlow}: Large-scale machine learning on heterogeneous systems,
  2015.
\newblock Software available from tensorflow.org.

\end{thebibliography}
\small



\section*{Checklist}


\begin{enumerate}

\item For all authors...
\begin{enumerate}
  \item Do the main claims made in the abstract and introduction accurately reflect the paper's contributions and scope?
    \answerYes{}
  \item Did you describe the limitations of your work?
    \answerYes{See Section~\ref{sec:conclusion}}
  \item Did you discuss any potential negative societal impacts of your work?
    \answerNA{}
  \item Have you read the ethics review guidelines and ensured that your paper conforms to them?
    \answerYes{}
\end{enumerate}

\item If you are including theoretical results...
\begin{enumerate}
  \item Did you state the full set of assumptions of all theoretical results?
    \answerNA{}
        \item Did you include complete proofs of all theoretical results?
    \answerNA{}
\end{enumerate}

\item If you ran experiments...
\begin{enumerate}
  \item Did you include the code, data, and instructions needed to reproduce the main experimental results (either in the supplemental material or as a URL)?
    \answerNo{The code and the data are proprietary. Whatever is in public domain is shared in the appendix.}
  \item Did you specify all the training details (e.g., data splits, hyperparameters, how they were chosen)?
    \answerYes{See section~\ref{sec:methods} and appendix~\ref{app:hypernetdetails}} 
            \item Did you report error bars (e.g., with respect to the random seed after running experiments multiple times)?
    \answerNA{}
        \item Did you include the total amount of compute and the type of resources used (e.g., type of GPUs, internal cluster, or cloud provider)?
    \answerYes{See appendix~\ref{app:hypernetdetails}}
\end{enumerate}

\item If you are using existing assets (e.g., code, data, models) or curating/releasing new assets...
\begin{enumerate}
  \item If your work uses existing assets, did you cite the creators?
    \answerNA{}
  \item Did you mention the license of the assets?
    \answerNA{}
  \item Did you include any new assets either in the supplemental material or as a URL?
    \answerNA{}
  \item Did you discuss whether and how consent was obtained from people whose data you're using/curating?
    \answerNA{}
  \item Did you discuss whether the data you are using/curating contains personally identifiable information or offensive content?
    \answerNA{}
\end{enumerate}

\item If you used crowdsourcing or conducted research with human subjects...
\begin{enumerate}
  \item Did you include the full text of instructions given to participants and screenshots, if applicable?
    \answerNA{}
  \item Did you describe any potential participant risks, with links to Institutional Review Board (IRB) approvals, if applicable?
    \answerNA{}
  \item Did you include the estimated hourly wage paid to participants and the total amount spent on participant compensation?
    \answerNA{}
\end{enumerate}

\end{enumerate}


\appendix

\section{Appendix}
\subsection{Heat Transfer in APH and the domain-decomposed PINN}
\label{app:domaindecomPINN}
\begin{figure}
    \centering
    \includegraphics[width=0.8\textwidth]{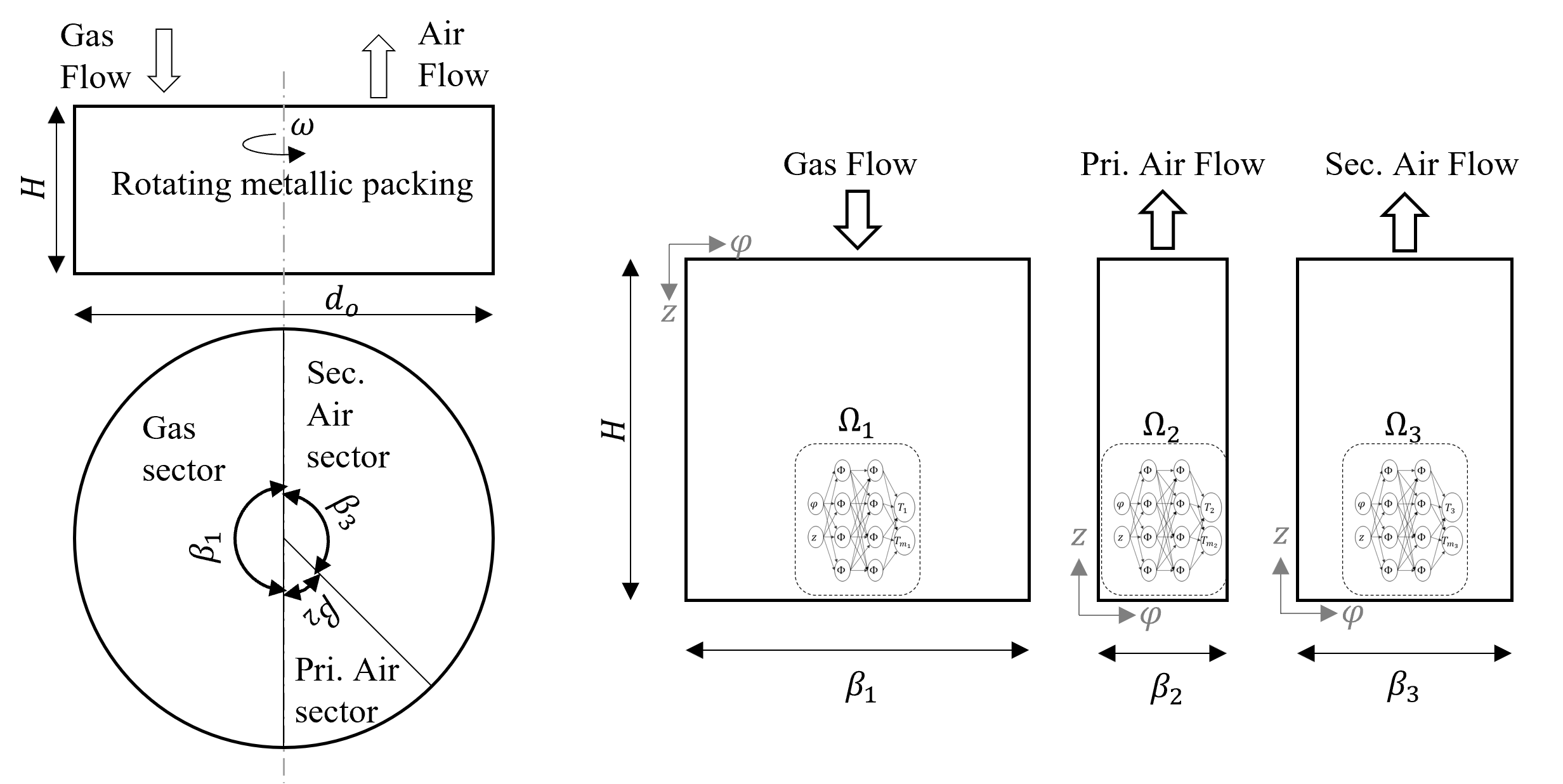}
    \caption{APH Schematic and computational domain \cite{Jadhav2022}}
    \label{fig:CompDomain}
\end{figure}

Figure \ref{fig:CompDomain} shows a schematic of APH and computational domain with co-ordinate system. Computational domain is divided into three sub-domains and each sub domain shares a common interface with the other two sub-domains. Subscript $j$ is used to represent these sub-domains. Equation \ref{eq:cond} and \ref{eq:conv} represent conduction and convection heat transfer in APH in non-dimensional form. Solution of these equations constitute the fluid temperature ($T$) and metal temperature ($T_m$) for given co-ordinates ($\varphi,z$). $NTU$ and $Pe$ represent the number of transfer units and Peclet number respectively.
\begin{equation}
\label{eq:cond}
    \frac{\partial T_{m_{j}}}{\partial\varphi} = NTU_{m_{j}} (T_j - T_{m_j}) + \frac{1}{Pe_{m_j}}   \frac{\partial^2 T_{m_j}}{\partial z^2} 
\end{equation}
\begin{equation}
\label{eq:conv}
    \frac{\partial T_j}{\partial z} = {NTU}_{m_j} \left( {T}_{m_{j}} - {T}_{j}, \right) \hspace{0.2cm} j = 1,2,3
\end{equation}
\begin{equation}
    \label{eq:bndcnd_gas}
    {T_j}\left( \varphi, z=0 \right) = {T}_{in,j},  \hspace{0.2cm} j = 1,2,3
\end{equation}
\begin{equation}
    \label{eq:intface_gas}
    T_{m_1}(\varphi = 0, z) = T_{m_3} (\varphi = 1, 1-z)
\end{equation}
\begin{equation}
    \label{eq:intface_priair}
    T_{m_1}(\varphi = 1, z) = T_{m_2} (\varphi = 0, 1 - z)
\end{equation}
\begin{equation}
    \label{eq:intface_secair}
    T_{m_2} (\varphi = 1, z) = T_{m_3} (\varphi = 0, z) 
\end{equation}
\begin{equation}
    \label{eq:gradcnd}
    \frac{\partial {T}_{m_j}[z = 0, 1]}{\partial z} = 0,  \hspace{0.2cm}j = 1,2,3 
\end{equation}
Temperature measurements for Gas inlet temperature ($T_{in,1}$), primary air inlet temperature ($T_{in,2}$), and secondary air inlet temperature ($T_{in,3}$) are used as boundary conditions (equation \ref{eq:bndcnd_gas}). 
Continuity constraints on the metal temperature due to rotation of the metal are given by equation \ref{eq:intface_gas},\ref{eq:intface_priair},\ref{eq:intface_secair}.
Constraint on the metal temperature gradient due to absence of axial conduction at the top and bottom of the subdomains is given by equation \ref{eq:gradcnd}.

The governing equations are solved using finite difference method \cite{Chung1983} and a grid independence study was performed to obtain the optimal grid size 240$\times$240, beyond which the predictions don't change significantly (see Table \ref{tab:grid_independence}). The results of finite difference method were validated with the experimental results provided in \cite{WANG201952}.

\begin{table}
  \caption{Grid Independence study of the physics model}
  \label{tab:grid_independence}
  \centering
  \begin{tabular}{ccc}
    \toprule

      {Grid Size (on gas and air side)} & $T_{2_{out}}$ (\degree C) & $T_{1_{out}}$(\degree C) \\
    \midrule
        30 $\times$ 30    &    354.93    &    126.35     \\
        60 $\times$ 60    &    357.26    &    123.60     \\
        120 $\times$ 120    &    358.39    &    122.25     \\
        240 $\times$ 240    &    358.93    &    121.58     \\
        480 $\times$ 480    &    359.20    &    121.26     \\

    \bottomrule
  \end{tabular}
\end{table}

\begin{figure}
    \centering
    \includegraphics[width=\textwidth]{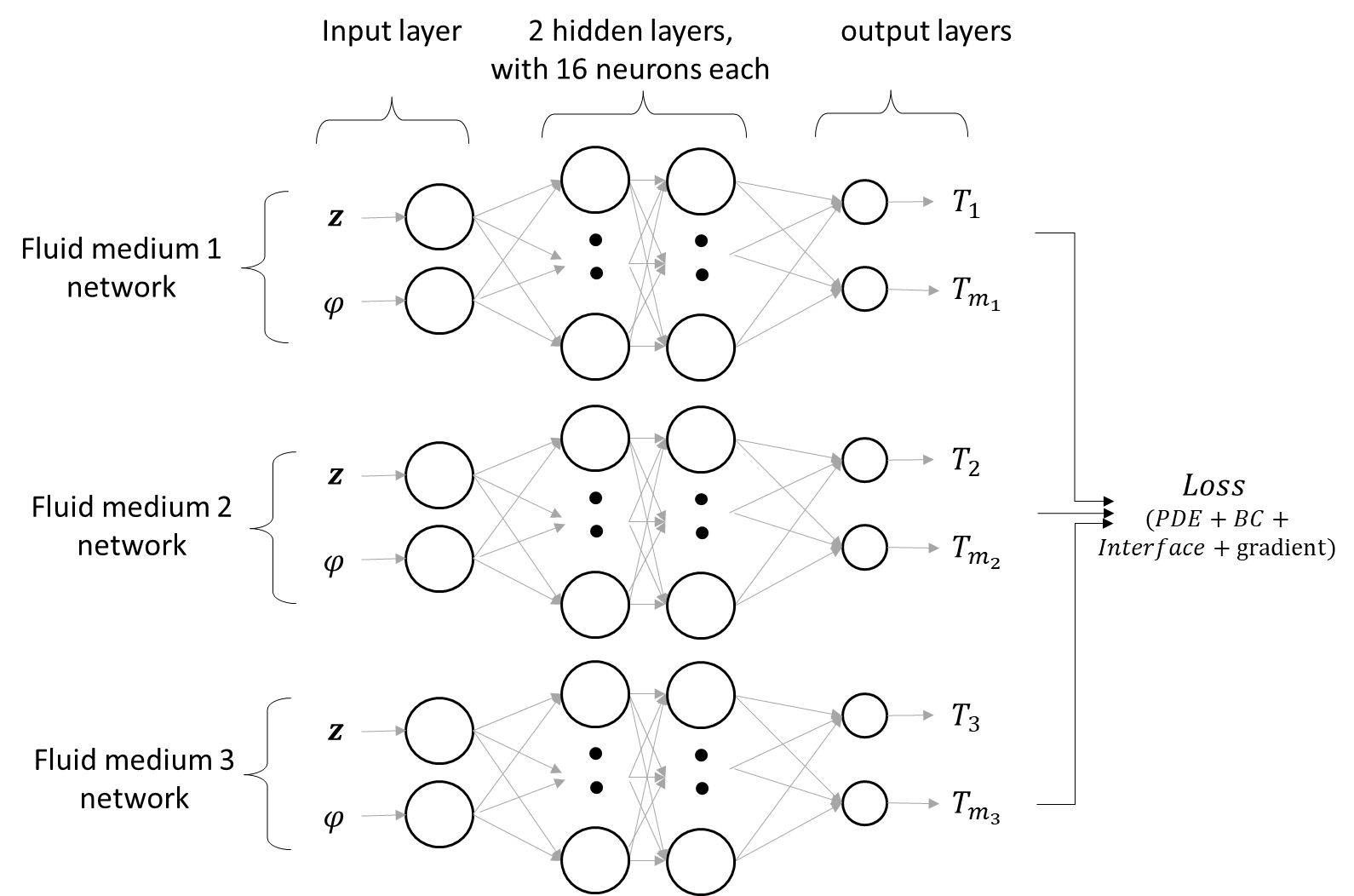}
    \caption{Domain Decomposed Base PINN architecture \cite{Jadhav2022}}
    \label{fig:domaindecompoPINN}
\end{figure}

Figure \ref{fig:domaindecompoPINN} shows the domain decomposed PINN \cite{Jadhav2022} used as base PINN in this work. Base PINN consists of three sub-networks which takes co-ordinates ($\varphi, z$) as input and outputs fluid and metal temperature. Each of the sub-network has input layer with 2 neurons, 2 hidden dense layers with 16 neurons and output layer with 2 neurons. Base PINN is trained using Adam optimizer. 
\subsection{Hypernetworks - additional details}
\label{app:hypernetdetails}
\begin{figure}
    \centering
    \includegraphics[width=\textwidth]{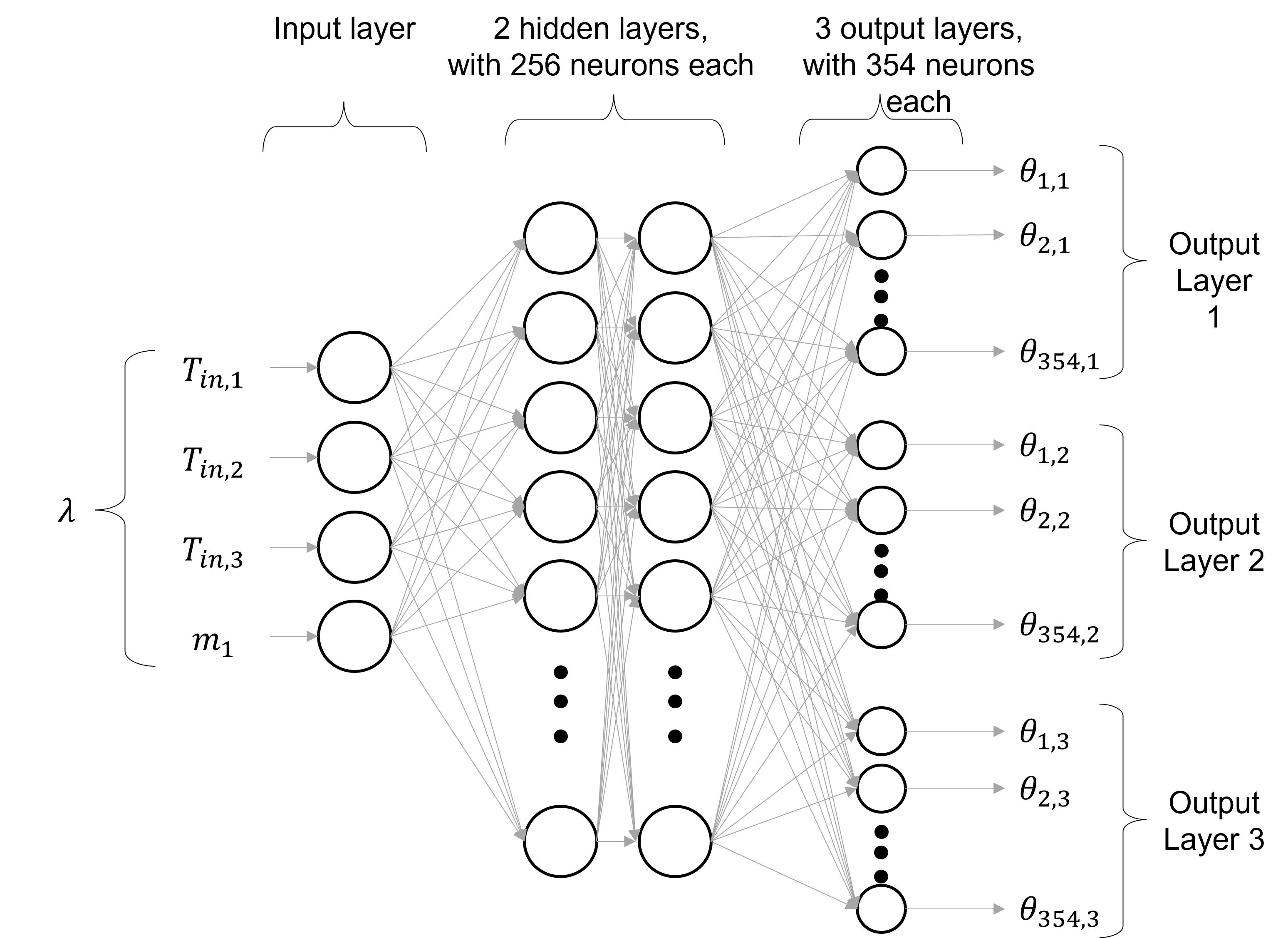}
    \caption{Hypernetwork Architecture}
    \label{fig:hypernetArch}
\end{figure}

Figure \ref{fig:hypernetArch} represents the architecture of hypernetwork, comprising one input layer (4 neurons), 2 hidden layers (256 neurons each and tanh activation) and 3 output layers (354 neurons each and linear activation). Hypernetwork is trained using Adam optimizer with a learning rate of 1e-4. Early stopping callback was used for stopping the training by monitoring the validation loss and a minimum delta threshold of 1e-6. Training and evaluation of networks was performed on a machine having an AMD Ryzen 5 5500U processor with Radeon graphics 2.10 GHz, 16 GB RAM, with Tensorflow \cite{tensorflow2015-whitepaper} as the deep learning framework.

The training tasks for the hypernetwork vary depending upon the total number of tasks in the design based on Taguchi method. The validation and the test tasks are given in table \ref{tab:validtesttask}

\begin{table}
  \caption{Maximum and minimum values considered for taguchi design}
  \label{tab:variablerange}
  \centering
  \begin{tabular}{lrr}
    \toprule
     & Minimum Value & Maximum Value \\
    \midrule
    Inlet Gas Temperature $T_{in,1}$ (\degree C) & 200 & 400     \\
    Inlet Primary Air Temperature $T_{in,2}$(\degree C) & 10 & 80      \\
    Inlet Secondary Air Temperature $T_{in,3}$(\degree C) & 10 & 80  \\
    Gas Flow Rate $m_1$(kg/s) & 600	& 800\\
    \bottomrule
  \end{tabular}
\end{table}

\begin{table}
  \caption{Validation and test tasks used for hypernetwork assessment}
  \label{tab:validtesttask}
  \centering
  \begin{tabular}{ccccc}
    \toprule
    \multicolumn{5}{c}{Validation Tasks}                   \\
    \cmidrule(r){1-5}
    task & $T_{in,1}$(\degree C) & $T_{in,2}$(\degree C) & $T_{in,2}$(\degree C) & $m_1$ (kg/s)\\
    \midrule
    1    &    206.36    &    45.01    &    39.56    &    769.01  \\
    2    &    215.47    &    66.02    &    60.47    &    627.49  \\
    3    &    216.36    &    38.01    &    29.11    &    668.11  \\
    4    &    234.15    &    62.99    &    57.68    &    697.88  \\
    5    &    242.47    &    23.35    &    18.18    &    680.32  \\
    6    &    253.50    &    52.02    &    47.27    &    623.54  \\
    7    &    259.12    &    32.91    &    29.26    &    628.07  \\
    8    &    281.47    &    40.60    &    31.34    &    769.81  \\
    9    &    290.43    &    49.47    &    41.01    &    619.54  \\
    10    &    305.69    &    64.71    &    55.12    &    689.05  \\
    11    &    318.63    &    47.02    &    39.80    &    779.24  \\
    12    &    333.94    &    49.59    &    43.96    &    694.81  \\
    13    &    340.12    &    24.60    &    18.84    &    751.40  \\
    14    &    367.38    &    58.36    &    53.24    &    639.89  \\
    15    &    373.33    &    15.55    &    11.50    &    649.93  \\
    16    &    379.90    &    24.18    &    15.05    &    741.61  \\
    17    &    386.80    &    26.87    &    22.95    &    700.05  \\
    18    &    387.55    &    70.64    &    65.04    &    742.06  \\
    19    &    403.56    &    33.74    &    29.93    &    611.67  \\
    \cmidrule(r){1-5}
    \multicolumn{5}{c}{Test Tasks}                   \\
    \cmidrule(r){1-5}
    task & $T_{in,1}$(\degree C) & $T_{in,2}$(\degree C) & $T_{in,2}$(\degree C) & $m_1$ (kg/s)\\
    \midrule
    1    &    220.76    &    64.23    &    56.86    &    632.42   \\
    2    &    228.96    &    36.18    &    31.26    &    733.69   \\
    3    &    238.17    &    66.39    &    60.71    &    637.70   \\
    4    &    239.28    &    42.42    &    37.12    &    719.83   \\
    5    &    259.04    &    20.87    &    14.56    &    766.32   \\
    6    &    260.89    &    55.96    &    46.17    &    661.23   \\
    7    &    280.41    &    46.87    &    42.31    &    709.90   \\
    8    &    282.55    &    58.91    &    50.03    &    642.26   \\
    9    &    289.06    &    22.44    &    15.81    &    615.61   \\
    10    &    300.00    &    21.59    &    16.69    &    790.06   \\
    11    &    301.80    &    55.02    &    50.10    &    622.88   \\
    12    &    307.74    &    23.99    &    14.24    &    684.91   \\
    13    &    314.51    &    47.60    &    39.18    &    680.84   \\
    14    &    339.71    &    67.90    &    62.03    &    784.34   \\
    15    &    341.12    &    28.50    &    23.07    &    758.78   \\
    16    &    365.33    &    59.99    &    56.01    &    606.63    \\
    17    &    386.19    &    58.01    &    50.42    &    768.40    \\
    18    &    391.75    &    68.53    &    62.12    &    701.84    \\
    19    &    398.48    &    53.51    &    49.94    &    746.56    \\
    \bottomrule
  \end{tabular}
\end{table}

\begin{figure}
    \centering
    \includegraphics[width=\textwidth]{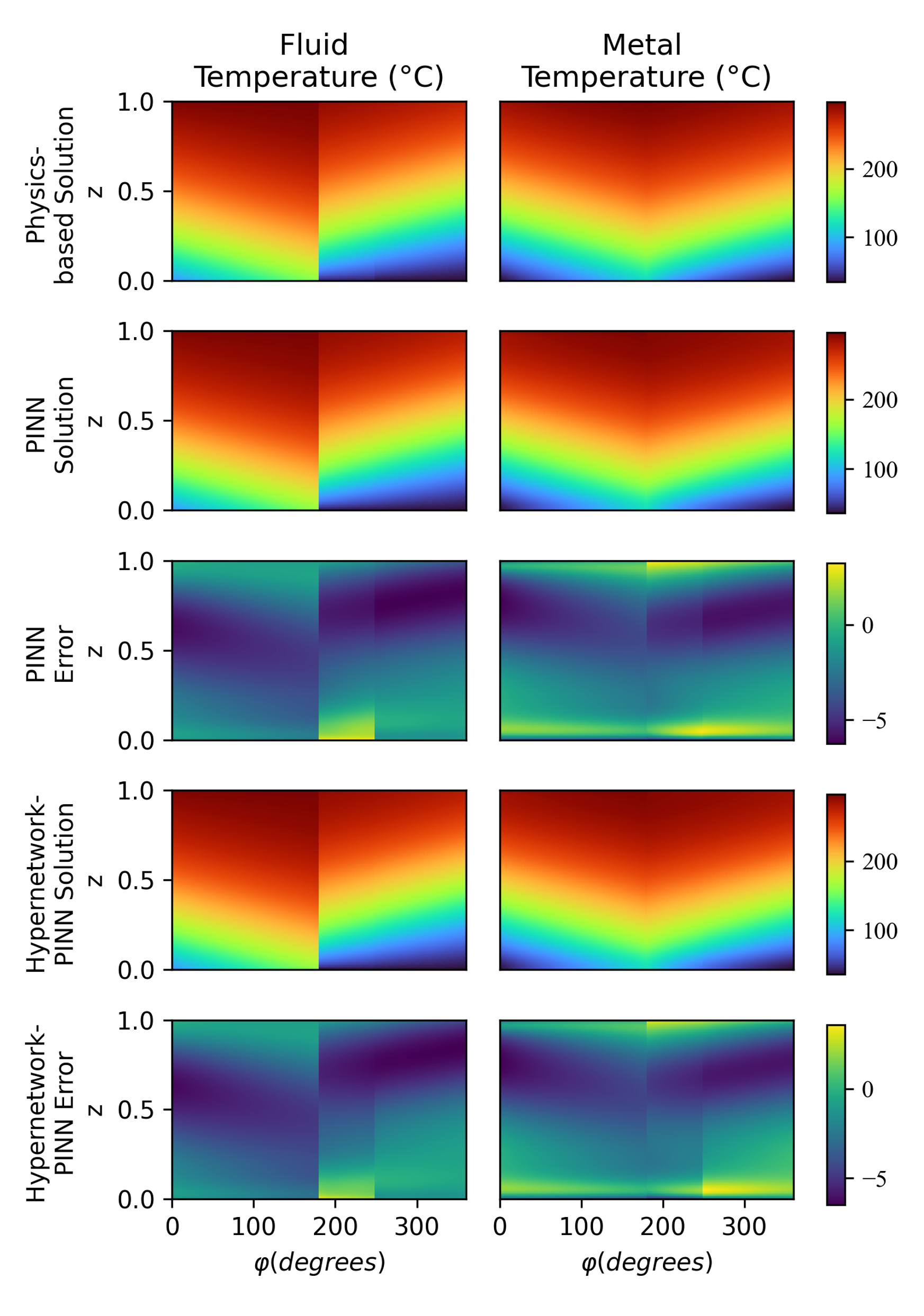}
    \caption{PINN, Hypernetwork and Physics Simulation Comparison}
    \label{fig:pinnhypernetworksoln}
\end{figure}

Figure \ref{fig:pinnhypernetworksoln} shows the comparison of thermal profile predicted with physics-based simulation, domain-decomposed PINN and the hypernetwork-based PINN for a representative test task ($T_{in,1}$ = 300.00 \degree C , $T_{in,2}$= 21.59 \degree C , $T_{in,2}$=16.69 \degree C , $m_1$=790.06 kg/s). 

\subsection{Taguchi Design of Experiments}
\label{app:Taguchi}
The designs proposed by Taguchi \cite{taguchi} comprise orthogonal arrays that structure the influential variables of a process for efficient experimentation. Therefore, instead of testing all the possible combinations of variables, the method provides a limited set of combinations that assists in identifying the most influential variables with minimal time and cost. Typically, standard designs (designated as L4, L16 and so on) for different number of variables and their levels are available and can be manipulated to create new experiment design as per the need. 

\end{document}